\pdfoutput=1

\documentclass[11pt]{article}

\usepackage[final]{acl}

\usepackage{times}
\usepackage{latexsym}

\usepackage[T1]{fontenc}

\usepackage[utf8]{inputenc}

\usepackage{microtype}
\usepackage{rotating} 

\usepackage{inconsolata}

\usepackage{amsthm, bbm}
\usepackage{amsfonts}
\usepackage{inconsolata}
\usepackage{subcaption}
\usepackage{booktabs}
\usepackage{xcolor}
\usepackage{comment}
\usepackage{hyperref}
\usepackage{tcolorbox}
\usepackage{dsfont,amssymb,amsmath, graphicx,enumitem}
\usepackage{pgfplotstable}
\usepackage{multirow}
\usepackage{booktabs}
\theoremstyle{definition}

\theoremstyle{remark}

\title{Structured Pruning for Diverse Best-of-$N$ Reasoning Optimization}

\author{Hieu Trung Nguyen, \qquad Bao Nguyen, \qquad Viet Anh Nguyen \\
  The Chinese University of Hong Kong \\
  \texttt{\{thnguyen, nbnguyen, nguyen\}@se.cuhk.edu.hk}}

\begin{document}
\maketitle
\begin{abstract}
Model pruning in transformer-based language models, traditionally viewed as a means of achieving computational savings, can enhance the model's reasoning capabilities. In this work, we uncover a surprising phenomenon: the selective pruning of certain attention heads leads to improvements in reasoning performance, particularly on challenging tasks. Motivated by this observation, we propose SPRINT, a novel contrastive learning framework that dynamically selects the optimal head and layer to prune during inference. By aligning question embeddings with head embeddings, SPRINT identifies those pruned-head configurations that result in more accurate reasoning. Extensive experiments demonstrate that our method significantly outperforms traditional best-of-$N$ and random head selection strategies on the MATH500 and GSM8K datasets. 
\end{abstract}

\section{Introduction}

Language models (LMs) have recently made remarkable progress in understanding and reasoning with human language, attracting considerable research attention~\cite{brown2020language, hoffmann2022training, chowdhery2022palm, rae2021scaling, raffel2019exploring}. This unprecedented evolution opened up a wide range of practical applications, including content generation~\cite{li2024pre, liu2023summary, achiam2023gpt}, virtual assistance~\cite{sezgin2024redefining, doe2023natural, garcia2023ethical}, population simulation~\cite{ref:bui2025mixture}, and problem solving~\cite{ahn2024large, imani2023mathprompter, lewkowycz2022solving}.

Despite these advances, the reliability and correctness of LM-generated answers remain a significant concern. Analyses of their output frequently reveal faulty reasoning and factual inaccuracies~\cite{ji2023survey,xu2024hallucination, nguyen2025risk, jiang2025probe, li2024inference}. This issue is particularly pronounced in tasks that require advanced reasoning, such as automated theorem proving~\cite{ref:wu2022autoformalization}, mathematical problem solving~\cite{ref:trinh2024solving}, or heuristic discovery~\cite{ref:romera2024mathematical}. LLMs often struggle to produce accurate responses in these scenarios in a single pass.

To address this challenge, iterative generation strategies have been employed to refine and select the most appropriate response, often by combining language generation with aggregation or search techniques such as best-of-N sampling~\cite{stiennon2020learning} or Monte Carlo Tree Search~\cite{ref:xie2024monte}. Throughout this paper, we refer to these iterative approaches as reasoning methods.

Current reasoning methods typically leverage the diversity induced by the LLM's decoding process through parameters such as temperature, top-$k$, and top-$p$ to generate multiple candidate answers~\cite{stiennon2020learning, ref:xie2024monte}. Although this technique introduces textual diversity and paraphrasing, it does not capture fundamentally different viewpoints, since all candidates are produced by the same underlying model with similar reasoning capabilities. One potential remedy is to employ a variety of LMs to generate answers; however, this approach is often computationally expensive and memory inefficient. 

At the same time, recent research on model pruning has shown that comparable performance can be achieved by selectively disabling certain heads in the transformer architecture~\cite{ma2023llm, wang2019structured}. Removing an attention head does not necessarily weaken the overall ability of the model because a head may introduce noise into the decision-making process, leading to \textit{in}correct answers. In fact, our empirical observations suggest that pruning certain heads can \textit{improve} the reasoning methods' performance on specific categories of questions; Section~\ref{sec:empirical_observation} discusses this phenomenon in more detail. This finding motivates our approach: starting from a single base model, we create a set of new models by pruning a different attention head in each, one at a time. This allows us to build a diverse pool of pruned models without significant additional computational or memory overhead.

\textbf{Contributions.} We begin by analyzing and providing new insights into the surprising phenomenon that pruning attention heads can improve the performance of the language model on the MATH benchmark. Motivated by these insights, we introduce the Structured PRunIng for diverSe reasoNing opTimization (SPRINT) framework. SPRINT first learns a set of embeddings for each attention head-layer pair using contrastive learning, guided by a novel Diversity-Promoted Contrastive Loss. This loss function encourages the question embedding to move closer to the embeddings of heads whose pruning leads to correct answers for that question. Finally, we leverage SPRINT to enhance best-of-$N$ sampling by pruning the top $N$ nearest head-layer configurations, determined by the distance between the learned embeddings and the question embedding.


Experiments on the MATH500 and GSM8K datasets show that our method outperforms traditional best-of-$N$ sampling approaches, achieving higher accuracy without introducing inference-time overhead.

Our paper unfolds as follows: Section~\ref{sec:related} discusses related work on reasoning in LMs. Section~\ref{sec:how_does} investigates the impact of pruning individual attention heads in transformer models, revealing that certain heads can be removed without performance loss and, in some cases, even improve accuracy. Section~\ref{sec:method} delineates our SPRINT framework for matching heads for pruning with input questions, and Section~\ref{sec:exp} presents the extensive numerical results of the mathematical reasoning task. 

\section{Related Works}
\label{sec:related}

A straightforward decoding process may struggle to produce accurate solutions for complex reasoning tasks. To address this, Self-Consistency (SC) improves reliability by generating multiple outputs and selecting the final response by majority voting~\cite{wang2022self}. A similar approach, best-of-N sampling, employs a reward model or function to select the response with the highest reward \cite{stiennon2020learning}. While both techniques enhance output quality, they also increase computational cost proportional to the number of samples.

To explore reasoning pathways more effectively, tree-search-based methods have been introduced, such as Tree-of-Thought \cite{yao2024tree}, Monte Carlo Tree Search (MCTS) \cite{feng2023alphazero, zhang2024rest, guan2025rstar}, and Forest-of-Thought \cite{bi2024forest}. \citet{ref:damani2024learning} highlights that tree-based search strategies are more effective in discovering correct solutions than parallel sampling, particularly for complex problems.

Beyond reasoning strategies, scaling inference-time computation also plays a crucial role in improving output quality. \citet{beeching2024scalingtesttimecompute} shows that the accuracy on the MATH-500 benchmark improves as the test-time computation (that is, the number of generations per problem) increases for methods such as best-of-N, beam search, and diverse verifier tree search (DVTS). Similarly, \citet{guan2025rstar} achieved an average accuracy of 53.3\% on the USA Math Olympiad (AIME) by conducting extensive MCTS rollouts.

Our work is related to recent research on interventions in the internal state of language models~\cite{ref:wu2024pyvene, ref:li2023inference, ref:nguyen2025multi, ref:liu2023context, ref:nguyen2025task}. Although prior approaches often focus on learning intervention vectors to add to model activations, instead we investigate the effect of excluding specific components of the internal state of the model.

\section{Does Pruning an Attention Head Improve LM's Performance?}
\label{sec:how_does}

\begin{figure*}[h]
    \centering
    \includegraphics[width=0.95\linewidth]{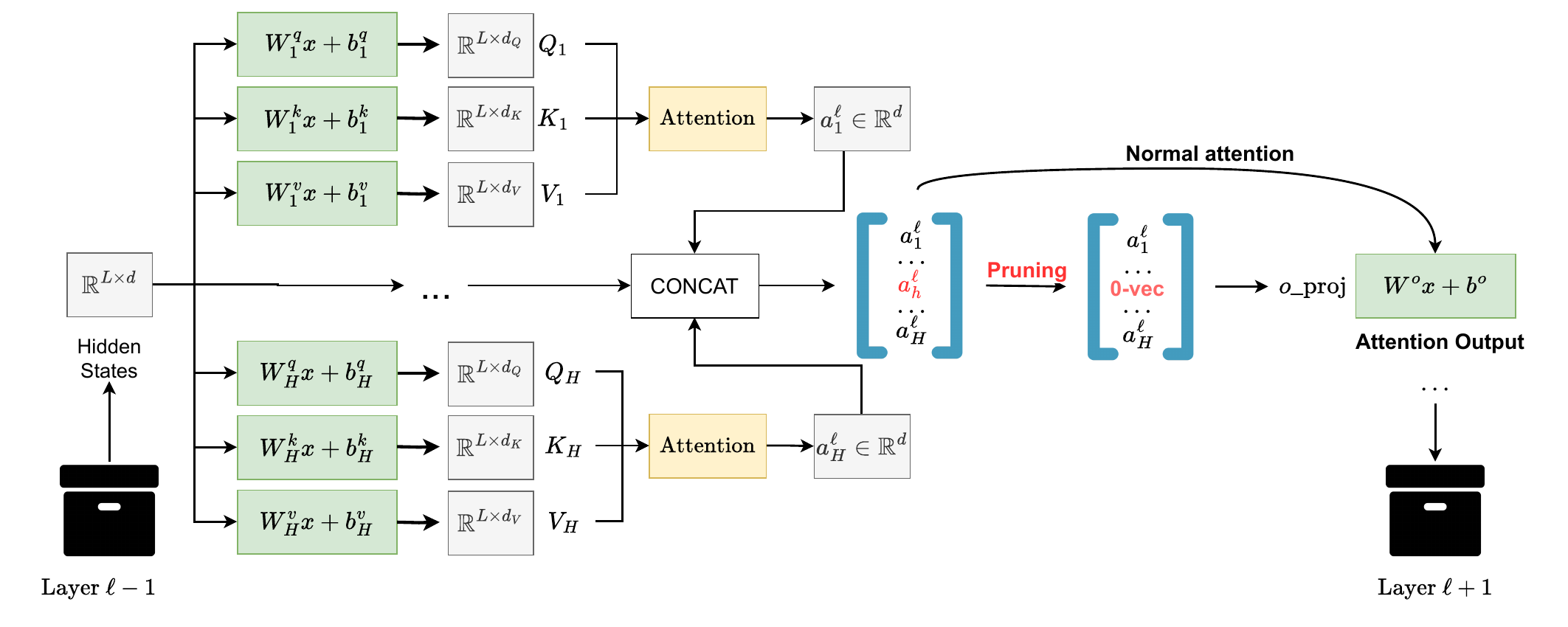}
    \caption{The figure illustrates our attention head pruning technique in a standard multi-head attention mechanism of transformer-based models. We define pruning the $h$-th attention head $a_h^\ell$ from layer $\ell$ by zeroing out its output vector before concatenating the head outputs and passing the result to the $\texttt{o}_\texttt{proj}$ layer.}
    \label{fig:pruning_illustration}
\end{figure*}

 We consider a transformer-based model, where each layer contains $H$ attention heads. In each layer $\ell$, the outputs of the $H$ attention heads are denoted as $\{a_h^\ell\}_{h=1}^{H}$, where each $a_h^\ell \in \mathbb{R}^{d}$. The concatenated vector, $a^\ell = (a_1^\ell, \ldots, a_H^\ell) \in \mathbb{R}^{D}$ with $D = H \times d$, is then passed to a linear layer to produce the output of a multi-head attention block.

\textbf{Attention head pruning.} In our study, we define \textbf{head pruning} as zeroing out the output vector $a_h^\ell$ of a specific attention head before the concatenation step, as illustrated in Figure~\ref{fig:pruning_illustration}. In modern decoder-only architectures~\cite{ref:llama3modelcard}, this is equivalent to pruning the input of $\texttt{o}_\texttt{proj}$ layer.

In this section, we detail our experimental procedure for attention head pruning and present a surprising empirical phenomenon that motivates our subsequent work.

\subsection{Experimental Procedure}

To systematically examine the effect of individual head pruning, we follow the procedure below:
\begin{enumerate}[leftmargin=5mm]
    \item \textbf{Layer and Head Selection:}  
    We select $L=4$ layers from the model, specifically, the first layer, layers 5 and 15, and the last layer, for a total of $LH$ attention heads. For each head, we create a variant of the base model with that specific head pruned, resulting in $LH$ model variants in addition to the base model. 

    \item \textbf{Measuring the Impact:}  
    For each model $m \in \mathcal{M}$, we generate solutions on the MATH500 dataset, which consists of $n=500$ samples. The predictions are represented by a binary accuracy vector $z \in \{0,1\}^{n}$, where $z_i = 1$ if sample $i$ is solved correctly and $0$ otherwise. Evaluating all models yields a matrix $Z \in \{0, 1\}^{n \times LH}$ of accuracy values.
\end{enumerate}

In this experiment\footnote{The project's repository is available at \url{https://github.com/HieuNT91/attention_pruning}.}, we study two models specialized for mathematical reasoning, \texttt{Qwen2.5-Math-1.5B-Instruct} and \texttt{Qwen2.5-Math-7B-Instruct}, as well as two general-purpose models, \texttt{Qwen2.5-7B-Instruct}, and \texttt{Meta-Llama-3-8B-Instruct}. Our experiments involve evaluating model performance with and without head pruning on 500 samples from the MATH500 dataset~\cite{ref:hendrycks2021measuring}.

\subsection{Empirical Observations}
\label{sec:empirical_observation}

Our empirical results, shown in Tables~\ref{tab:observe_llama}, \ref{tab:observe_qwen_math_1.5}, \ref{tab:observe_qwen_math_7}, \ref{tab:observe_qwen_7}, reveal several surprising findings about attention head pruning. Contrary to the common belief that all attention heads are essential for optimal model performance, we find that pruning individual heads often does not reduce accuracy, and in many cases, increases it. The impact of pruning depends strongly on both the specific head and its layer: while some heads are irreplaceable, for example, removing head 3 in layer 0 (see tables~\ref{tab:observe_qwen_math_1.5}, \ref{tab:observe_qwen_math_7}, and 
\ref{tab:observe_qwen_7}) causes performance to drop to near zero. This pattern mirrors findings from~\cite{ref:yu2024super}, which show that certain `superweights' in the early layers play a disproportionate role in overall LLM performance. Conversely, pruning less critical heads can preserve or even enhance performance. For instance, in the `Counting \& Probability' category, pruning head 0 in layer 15 of \texttt{Qwen2.5-Math-1.5B-Instruct} increased accuracy by 16\% over the unpruned baseline.

\begin{figure}[ht]
    \centering
    \includegraphics[width=1\linewidth]{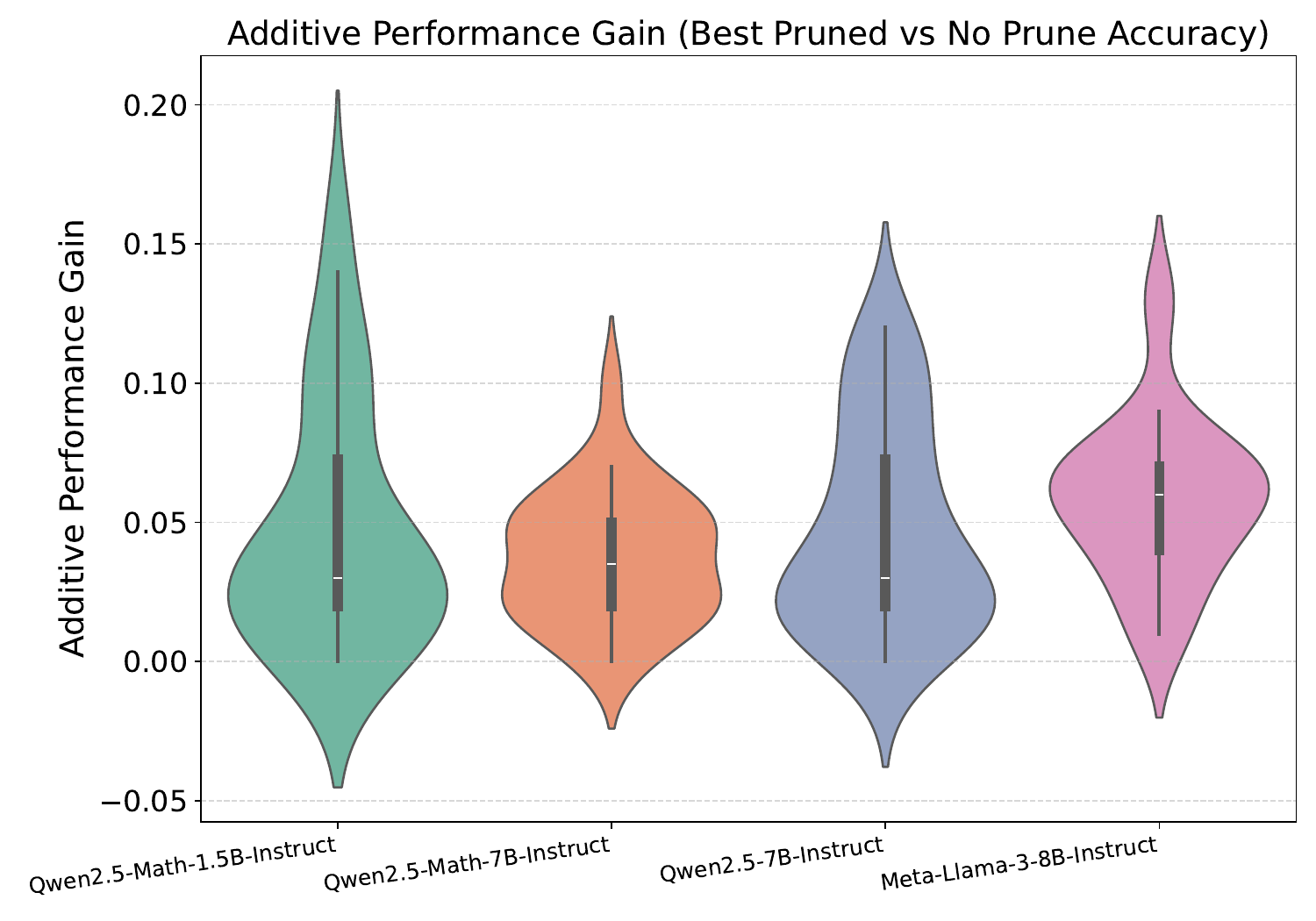}
    \caption{Distribution of additive performance gains (best pruned accuracy minus no-prune accuracy) for Qwen2.5-Math-1.5B-Instruct, Qwen2.5-Math-7B-Instruct, Qwen2.5-7B-Instruct, and Meta-Llama-3-8B-Instruct. Each violin illustrates the improvement across all subject-layer pairs.}
    \label{fig:performance_gain}
\end{figure}

We further illustrate this trend in Figure~\ref{fig:performance_gain}, which displays violin plots of additive performance gains (the difference between the best pruned head accuracy and the baseline accuracy) across all subcategories for each model. These plots show that head pruning consistently produces positive gains for all evaluated models. The specialized mathematical models exhibit higher median gains and more frequent large improvements, particularly in domains such as `Counting \& Probability' and `Geometry'. In contrast, the general-purpose models show more modest but still positive and consistent benefits from pruning.

Overall, these patterns suggest that some attention heads contribute redundant or even harmful information, and that pruning can act as a useful form of regularization. Our findings highlight the significant redundancy present in multi-head attention and indicate promising directions for simplifying models and improving efficiency through targeted head pruning.

\section{Methodology}
\label{sec:method}

\begin{figure*}[ht]
    \centering
    \includegraphics[width=0.8\linewidth]{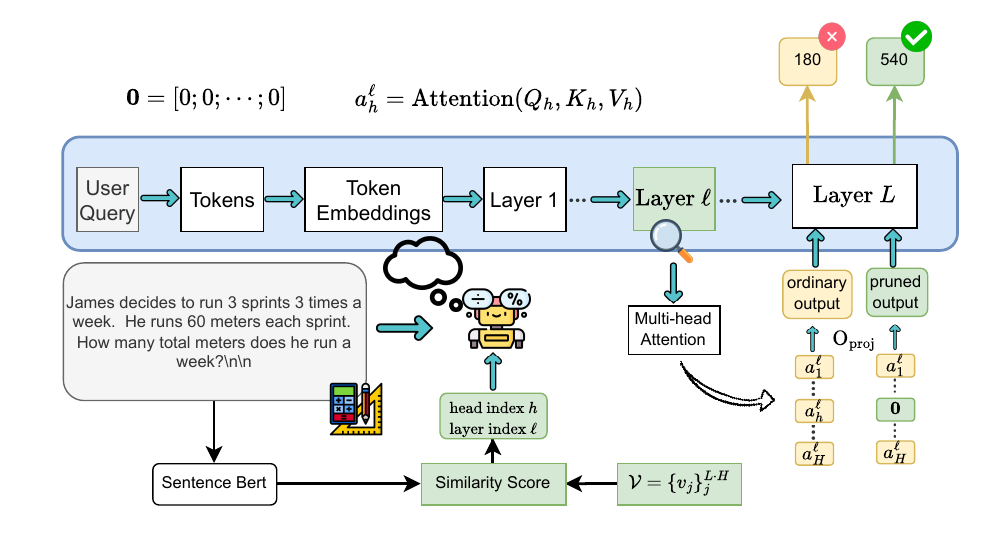}
    \caption{Schematic overview of our head pruning framework. First, we use contrastive learning to learn a set of embeddings $\mathcal{V}$, where each embedding represents a head-layer tuple ($h, \ell$). We then choose which head to prune by picking the embedding that minimizes the Euclidean distance of question $q_i$.}
    \label{fig:intro}
\end{figure*}

Motivated by these counterintuitive findings, we propose a framework called Structured Pruning for Diverse Reasoning Optimization (SPRINT) to leverage selective head pruning to improve reasoning performance. At the core of SPRINT is a contrastive learning approach that aligns question embeddings with attention head embeddings to identify the heads most beneficial to prune. Initially, every head is associated with a learnable vector from a set
$\mathcal{V} = \{v_j\}_{j=1}^{L H}$,
where each $v_j \in \mathbb{R}^p$ is randomly initialized. A sentence embedding model, denoted by $\phi$, processes each question $x_i$, after which a linear transformation $\theta$ produces the question embedding
$q_i = \theta\big(\phi(x_i)\big)$.
For each question, we define the set of pruned-head models as $\mathcal{M} = \{1, \ldots, L H\}$ and partition this set into a positive subset $\mathcal{M}_i^+ = \{j: z_{ij} = 1\}$ containing heads that result in correct answers to question $i$, and a negative subset $\mathcal{M}_i^- = \{j: z_{ij} = 0\}$.

To quantify the correspondence between different pruned-head models, we introduce a similarity measure between any two heads $j$ and $k$ as follows:
\[
s_{jk} = \frac{1}{n} \sum_{i=1}^{n} \mathbbm{1}(z_{ij} = z_{ik}) \in [0, 1],
\]
which computes the fraction of questions for which both heads produce the same outcome. We jointly optimize the transformation $\theta$ and the head embeddings $\mathcal{V}$ with the following loss function:
\begin{equation}
\begin{aligned}
\mathcal{L} = \; & \frac{1}{n}\sum_{i=1}^{n} -\log \left( \frac{\sum_{j \in \mathcal{M}_i^+} \exp\Big(-\|q_i - v_j\|_2^2\Big)}{\sum_{j'=1}^{L H} \exp\Big(-\|q_i - v_{j'}\|_2^2\Big)} \right) \\
& \quad - \lambda \sum_{j < k} s_{jk}\, \|v_j - v_k\|_2^2. \label{eq:loss1}
\end{aligned}
\end{equation}

Our joint optimization loss $\mathcal L$ consists of two components. The first term aims to align the question embedding \(q_i\) with the embeddings \(v_j\) of those pruned-head models that yield the correct answer. The loss employs a softmax formulation for each question, over the negative squared Euclidean distances between \(q_i\) and the head embeddings. This loss encourages the model to produce a lower distance, and thus a higher similarity, for those heads in the positive set \(\mathcal{M}_i^+\). 

The second term of the loss addresses the need for diversity among head embeddings. By summing the pairwise distances weighted by the similarity measure \(s_{jk}\), it acts as a regularizer that encourages distinctiveness between head embeddings, especially for those heads that tend to yield similar outcomes. This term effectively pushes similar heads apart in the embedding space to reduce redundancy and promote a richer, more discriminative representation. Overall, combining these two terms helps align the question representation with high-performing pruned models while maintaining a diverse set of head embeddings.

\textbf{Inference.} At inference time, we obtain the embedding for the new question $x_i$ by feedforwarding $q_i = \theta (\phi(x_i))$. The optimal head to prune for a given test question is identified by selecting the head whose embedding minimizes the squared Euclidean distance to the question embedding, i.e.,
\[
j^\star = \arg\min_{j \in \mathcal{M}} \|q_i - v_j\|_2^2.
\]
We integrate best-of-$N$ reasoning into our framework to enhance the performance by selecting the top-$N$ pruned-head configurations based on their proximity to $q_i$. We obtain candidate answers via greedy decoding with a temperature of 0. 

Overall, our unified methodology demonstrates that pruning the attention heads, guided by a contrastive alignment strategy, can improve model performance. Thus, SPRINT identifies which heads benefit from pruning and supports a dynamic simulation approach to generate optimal predictions.

\section{Numerical Experiments}
\label{sec:exp}

\begin{figure}[ht]
    \centering
    \includegraphics[width=1.0\linewidth]{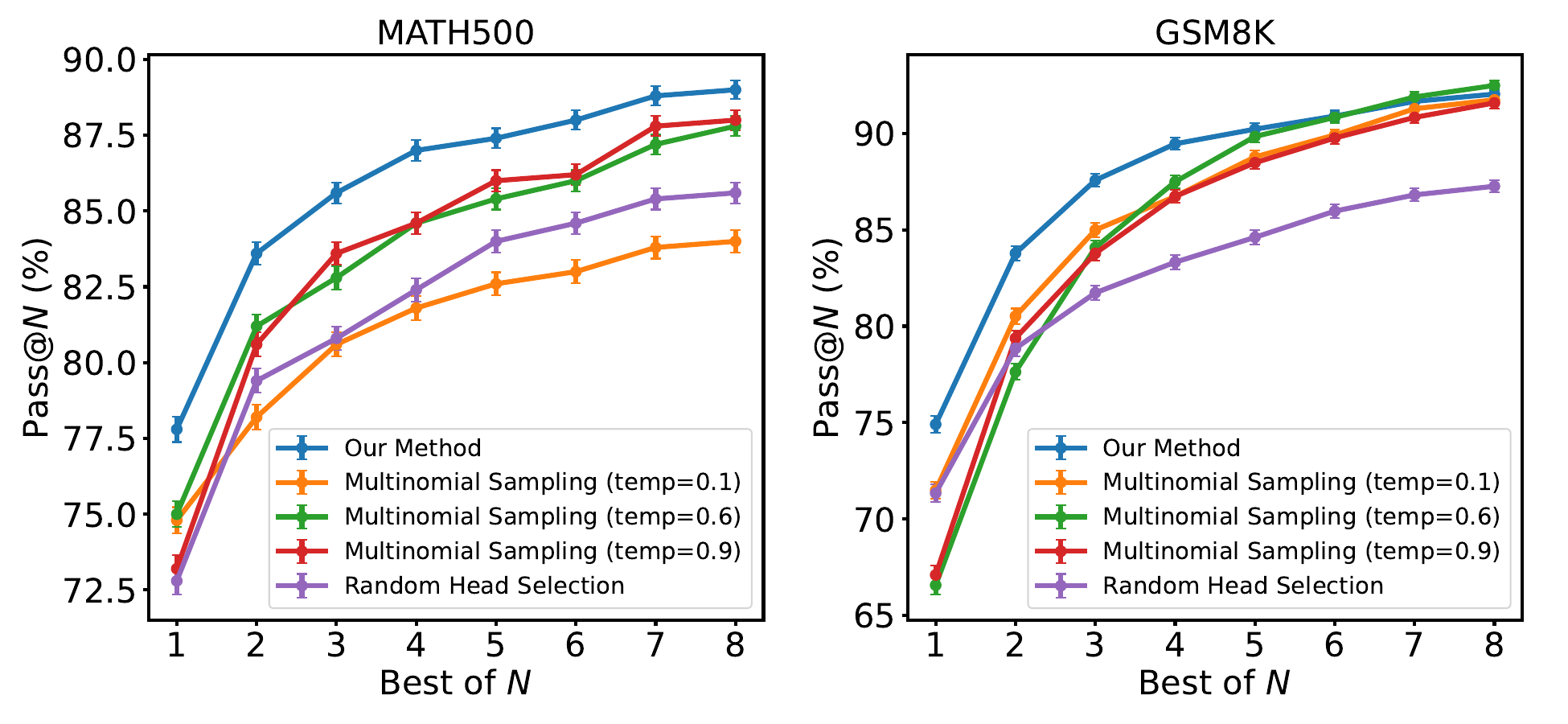}
    \caption{Average Pass@$N$ on MATH500 and GSM8K dataset.}
    \label{fig:main_results}
\end{figure}

\noindent \textbf{Datasets.} For our experiments, we use the MATH and GSM8K datasets. We provide more details in the appendix. 

\noindent \textbf{Base Models.}
As a proof of concept, we select \texttt{Qwen2.5-Math-1.5B-Instruct} as our generation model. This choice is motivated by its strong empirical gains observed in Figure~\ref{fig:performance_gain}. All experiments are performed using the original, unquantized version of this model. Due to resource constraints, we consider five layers \{5, 10, 15, 20, 25\} for pruning.

\noindent \textbf{Performance metrics.} To evaluate performance, we use the Pass@$N$ metric. For each method, we sample $N$ candidate answers and compute Pass@$N$ based on whether at least one candidate is correct, as determined by an oracle.

\noindent \textbf{Baselines.} We compare our methods against two baselines: multinomial sampling, commonly used for Best-of-$N$ Reasoning, and a random head selection strategy. We perform Best-of-$N$ evaluations using three temperature values for multinomial sampling: 0.1, 0.6, and 0.9. For the random head selection strategy, we use a greedy approach to identify the heads that solve the most samples in the training set. During testing, we randomly select $N$ heads and measure Pass@$N$.

\subsection{Quantitative Results}

Figure~\ref{fig:main_results} summarizes our primary findings. Our method consistently outperforms the baselines for nearly all values of $N$ across the MATH and GSM8K datasets. Notably, the greatest improvements are observed at lower values $N\in \{1,2,3,4\}$, while the performance gains become more gradual as $N>4$.


\section{Conclusion}
In this work, we demonstrated that pruning attention heads, a technique traditionally viewed as harmful, can enhance the reasoning capabilities of transformer-based language models. Our study revealed that the selective removal of specific heads mitigates the propagation of redundant or noisy signals, thereby enhancing overall model performance. We developed a contrastive learning framework to harness this phenomenon that dynamically selects the optimal attention head-layer configurations to prune at inference time. Experimental results confirm that our approach significantly outperforms traditional Best-of-$N$ methods and random head selection strategies. 

\newpage
\section*{Limitations}
Due to the space limitations of a short paper, our work does not experiment with a more diverse range of model architectures and sizes. Our work primarily focuses on the mathematical domain, but it can be extended to other scientific domains that require large language models' reasoning. Furthermore, we did not study an interesting case where pruning a single head leads to a catastrophic performance drop.

\bibliography{bib}

\appendix

\section{Implementation details.}
\label{sec:appendix}

\textbf{Datasets.} The MATH dataset contains 12,500 math problems, with 12,000 problems designated for training and 500 for testing~\cite{ref:lightman2023lets}. Similarly, the GSM8K dataset~\cite{ref:cobbe2021gsm8k} comprises 8,792 math problems, with 7,473 used for training and 1,319 for testing. In our experiments, we subsample 1,500 training examples from the respective training sets. The attention head embeddings $\mathcal{V}$ are then trained using the subsampled training set. Finally, we evaluate performance on the test splits of the aforementioned datasets.

\section{Empirical Observation on the Impact of Attention Head Pruning.}

\begin{figure}[h]
    \centering
    \includegraphics[width=1.2\linewidth]{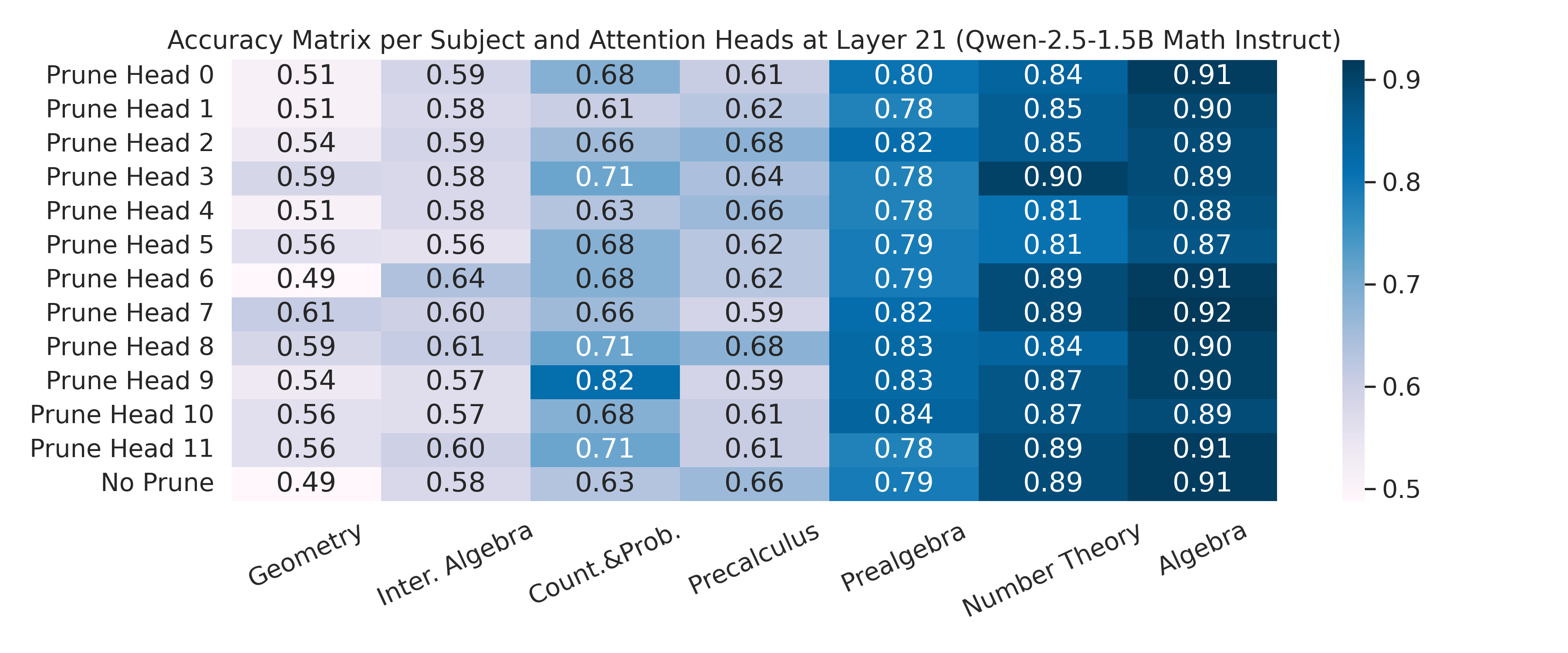}
    \caption{Impact of individual head pruning on model performance.}
    \label{fig:attention_head_pruning}
\end{figure}

Figure~\ref{fig:attention_head_pruning} shows that pruning an attention head in \texttt{Qwen-2.5-1.5B-Math-Instruct} model mostly leads to improvement in accuracy on the MATH500 dataset.

\clearpage
\begin{sidewaystable}
    \vspace{60mm}
    \begin{center}
    \caption{We report Pass@1 scores for single head pruning across different layers of \texttt{Meta-Llama-3-8B-Instruct}, organized by question statement subjects.}
    \label{tab:observe_llama}
    \resizebox{\textwidth}{!}{
    \begin{filecontents*}{llama8b.csv}
Method,Algebra,Algebra,Algebra,Algebra,Counting & Probability,Counting & Probability,Counting & Probability,Counting & Probability,Geometry,Geometry,Geometry,Geometry,Intermediate Algebra,Intermediate Algebra,Intermediate Algebra,Intermediate Algebra,Number Theory,Number Theory,Number Theory,Number Theory,Prealgebra,Prealgebra,Prealgebra,Prealgebra,Precalculus,Precalculus,Precalculus,Precalculus
Layer,5,10,15,31,5,10,15,31,5,10,15,31,5,10,15,31,5,10,15,31,5,10,15,31,5,10,15,31
Prune Head 0,0.35,0.37,0.32,0.32,0.11,0.11,0.13,0.16,0.2,0.24,0.2,0.22,0.13,0.14,0.14,0.16,0.19,0.19,0.21,0.18,0.4,0.38,0.41,0.39,0.14,0.12,0.14,0.12
Prune Head 1,0.31,0.37,0.32,0.34,0.11,0.11,0.08,0.08,0.2,0.22,0.22,0.24,0.14,0.13,0.14,0.13,0.19,0.24,0.19,0.24,0.41,0.37,0.37,0.4,0.11,0.14,0.11,0.11
Prune Head 2,0.33,0.32,0.32,0.34,0.16,0.13,0.11,0.11,0.22,0.15,0.22,0.2,0.14,0.13,0.15,0.15,\textbf{0.27},0.18,0.24,0.18,0.34,0.34,0.41,0.41,0.11,0.09,0.09,0.12
Prune Head 3,0.35,0.31,0.34,0.33,0.08,0.11,0.11,0.13,0.17,0.22,0.24,0.22,0.13,0.1,\textbf{0.16},0.16,0.19,0.16,0.21,0.19,0.38,\textbf{0.48},0.35,0.34,0.09,\textbf{0.18},0.11,0.09
Prune Head 4,0.31,0.34,0.3,0.33,0.11,0.11,0.16,0.11,0.22,0.22,0.22,0.22,0.14,0.13,0.15,0.14,0.15,0.21,0.23,0.19,\textbf{0.45},0.39,0.44,0.38,0.12,0.07,0.11,0.11
Prune Head 5,0.33,0.33,0.36,0.33,0.16,0.08,0.13,0.11,0.2,0.24,0.24,0.24,0.12,0.12,0.1,0.11,0.16,\textbf{0.27},0.23,0.21,0.39,0.37,0.4,\textbf{0.43},0.11,0.12,0.11,0.09
Prune Head 6,0.35,0.35,0.38,0.32,0.13,0.11,0.13,0.13,0.2,0.2,0.22,0.24,0.1,0.13,0.1,0.14,0.26,0.18,0.23,0.19,0.38,0.43,0.39,0.39,0.09,\textbf{0.18},0.12,0.11
Prune Head 7,0.35,0.34,0.31,0.31,0.11,0.05,0.13,0.13,0.17,0.24,0.17,0.22,0.11,0.11,0.12,0.16,0.21,0.21,0.21,0.19,0.39,0.4,0.43,0.35,0.07,0.09,0.14,0.11
Prune Head 8,0.35,0.37,0.31,0.34,0.13,0.18,0.11,0.16,0.22,0.24,0.22,0.2,0.11,0.12,0.12,0.13,0.23,0.21,0.13,0.21,0.41,0.38,0.34,0.4,0.12,0.12,\textbf{0.18},0.11
Prune Head 9,0.35,0.31,0.36,0.32,0.13,0.08,0.08,0.13,0.22,0.17,0.24,0.22,0.13,0.1,0.13,0.16,0.21,0.23,\textbf{0.26},0.21,0.41,0.39,0.34,0.39,0.07,0.11,0.16,0.11
Prune Head 10,0.33,0.37,0.38,0.34,0.16,0.08,0.08,0.13,0.24,0.2,0.17,0.24,0.14,0.09,0.14,0.15,0.19,0.21,0.21,0.21,0.4,0.4,0.39,0.4,0.11,\textbf{0.18},0.12,0.12
Prune Head 11,0.33,0.34,0.32,0.33,0.11,0.05,\textbf{0.18},0.13,0.22,0.24,0.27,0.2,\textbf{0.16},0.12,0.13,0.16,0.21,0.18,0.18,0.19,0.37,0.38,0.37,0.37,0.14,0.12,0.12,0.11
Prune Head 12,0.35,0.34,0.33,0.35,0.11,0.13,\textbf{0.18},0.13,0.27,0.22,0.2,0.24,0.14,0.12,0.14,0.12,0.24,0.19,0.19,0.21,0.38,0.44,0.43,0.39,0.12,0.12,0.12,0.07
Prune Head 13,0.35,0.35,0.27,0.35,0.13,0.05,\textbf{0.18},\textbf{0.18},0.15,0.24,0.1,0.2,\textbf{0.16},0.12,0.11,0.14,0.23,0.21,0.16,0.21,0.41,0.41,0.28,\textbf{0.43},0.11,0.12,0.12,0.11
Prune Head 14,0.35,0.38,0.32,0.27,0.11,\textbf{0.24},0.11,0.08,0.2,0.2,0.17,0.17,0.14,0.13,0.13,0.12,0.23,0.23,0.16,0.18,0.38,0.38,0.3,0.37,0.12,0.12,0.11,\textbf{0.14}
Prune Head 15,0.3,0.31,0.35,0.32,0.13,0.11,0.16,0.16,0.15,0.17,0.2,0.22,0.1,\textbf{0.16},0.09,0.13,0.23,0.21,0.23,0.19,0.41,0.35,0.35,0.41,0.14,0.11,0.11,0.07
Prune Head 16,0.35,0.35,0.33,0.33,0.13,0.11,0.13,0.13,0.22,0.24,0.22,0.24,0.13,0.14,0.15,0.15,0.18,0.18,0.21,0.21,0.39,0.4,0.44,0.39,\textbf{0.18},0.11,0.12,0.11
Prune Head 17,0.33,0.36,0.35,0.35,0.13,0.13,0.13,0.11,0.17,0.2,0.24,0.22,0.12,0.11,0.14,0.15,0.13,0.18,0.21,0.16,0.34,0.37,0.39,0.4,0.12,0.12,0.12,0.11
Prune Head 18,0.35,0.34,0.36,0.36,0.11,0.13,0.05,0.08,0.2,0.2,\textbf{0.29},0.17,0.14,0.12,0.11,0.15,0.24,0.24,0.23,0.23,0.35,0.37,0.41,0.37,0.16,0.14,0.11,0.11
Prune Head 19,0.32,0.36,0.34,0.33,0.11,0.13,0.13,0.13,0.22,0.22,0.2,\textbf{0.27},0.1,0.12,0.11,0.15,0.16,0.21,0.19,0.19,0.35,0.37,0.39,0.35,0.07,0.16,0.12,0.11
Prune Head 20,0.31,0.37,0.34,0.35,0.13,0.11,0.05,0.16,0.22,0.24,0.22,0.2,0.12,0.14,0.12,0.15,0.21,0.21,0.16,0.21,0.38,0.37,0.37,0.38,0.11,0.09,0.11,0.12
Prune Head 21,0.32,0.32,0.33,0.34,0.13,0.13,0.16,0.11,0.24,0.15,0.24,0.24,0.14,0.09,\textbf{0.16},0.15,0.23,0.11,0.16,0.21,0.4,0.38,0.39,0.38,0.11,0.14,0.11,0.11
Prune Head 22,0.35,0.37,0.35,0.35,0.05,0.13,0.16,0.08,0.22,0.2,0.2,0.24,0.14,0.14,\textbf{0.16},0.15,0.15,0.19,0.21,0.21,0.38,0.4,0.39,0.39,0.12,0.14,0.14,0.11
Prune Head 23,0.35,\textbf{0.40},0.33,0.35,0.13,0.13,0.11,0.16,0.2,0.22,0.22,0.22,0.15,0.13,0.13,\textbf{0.18},0.23,0.18,0.23,0.23,0.39,0.38,0.41,0.4,0.11,0.14,0.12,0.11
Prune Head 24,0.36,0.33,\textbf{0.39},0.32,\textbf{0.21},0.13,0.05,0.08,\textbf{0.29},0.27,0.2,0.1,0.12,0.14,0.14,0.13,0.18,0.24,0.18,\textbf{0.29},0.38,0.35,0.4,0.41,0.11,0.11,0.09,0.12
Prune Head 25,0.34,0.32,0.33,0.34,0.11,0.13,0.08,0.11,0.22,0.2,0.2,0.2,0.09,0.13,0.12,0.14,0.21,0.21,0.18,0.21,0.38,0.44,\textbf{0.46},\textbf{0.43},0.11,0.11,0.12,0.11
Prune Head 26,0.32,0.27,0.35,\textbf{0.40},0.11,0.16,0.13,0.08,0.2,0.24,0.2,0.17,0.14,0.12,0.12,0.14,0.18,0.18,0.16,0.23,0.4,0.41,0.39,0.4,0.12,0.14,0.14,0.12
Prune Head 27,0.34,0.31,0.33,0.34,0.08,0.11,0.08,0.13,0.22,\textbf{0.29},0.24,\textbf{0.27},0.14,0.14,\textbf{0.16},0.14,0.19,0.26,0.18,0.26,0.38,0.35,0.41,0.39,0.11,0.09,0.09,0.07
Prune Head 28,0.32,0.31,0.35,0.35,0.11,0.21,0.13,0.13,0.2,0.22,0.24,0.2,0.11,0.1,0.15,0.14,0.21,0.19,0.16,0.21,0.38,0.4,0.39,0.38,0.12,0.12,0.11,0.09
Prune Head 29,\textbf{0.37},0.38,0.34,0.33,0.13,0.13,0.08,0.11,\textbf{0.29},0.2,0.22,0.24,0.14,0.15,0.15,0.15,0.21,0.16,0.19,0.21,0.41,0.43,0.39,0.4,0.07,0.14,0.14,0.11
Prune Head 30,0.35,0.34,0.35,0.33,0.13,0.16,0.16,0.05,0.2,0.17,0.24,0.24,0.13,0.13,0.14,0.15,0.19,0.23,0.19,0.21,0.39,0.38,0.37,0.38,0.11,0.14,0.16,0.11
Prune Head 31,0.31,0.33,0.37,0.33,0.13,0.13,0.16,0.16,0.27,0.24,0.2,0.24,0.14,0.12,0.13,0.14,0.19,0.16,0.23,0.19,0.4,0.4,0.39,0.39,0.11,0.14,0.11,0.11
No Prune,0.33,0.33,0.33,0.33,0.11,0.11,0.11,0.11,0.24,0.24,0.24,0.24,0.15,0.15,0.15,0.15,0.21,0.21,0.21,0.21,0.39,0.39,0.39,0.39,0.11,0.11,0.11,0.11
\end{filecontents*}
\pgfplotstabletypeset[
        col sep=comma,
        string type,
        every head row/.style={
            output empty row,
            before row={
                \toprule
                \multirow{3}{*}{} &
                \multicolumn{4}{c}{Algebra} &
                \multicolumn{4}{c}{Counting \& Probability} &
                \multicolumn{4}{c}{Geometry} &
                \multicolumn{4}{c}{Intermediate Algebra} &
                \multicolumn{4}{c}{Number Theory} &
                \multicolumn{4}{c}{Prealgebra} &
                \multicolumn{4}{c}{Precalculus}\\
                \cmidrule(r){2-5}\cmidrule(r){6-9}\cmidrule(r){10-13}
                \cmidrule(r){14-17}\cmidrule(r){18-21}\cmidrule(r){22-25}
                \cmidrule(r){26-29}\\
            }
        },
        every row no 0/.style={after row=\midrule},
        every last row/.style={after row=\bottomrule},
    ]{llama8b.csv} 
    }
    \end{center}
\end{sidewaystable}
\clearpage

\clearpage
\begin{sidewaystable}
    \vspace{60mm}
    \begin{center}
    \caption{We report Pass@1 scores for single head pruning across different layers of \texttt{Qwen2.5-Math-1.5B-Instruct}, organized by question statement subjects.}
    \label{tab:observe_qwen_math_1.5}
    \resizebox{\textwidth}{!}{
    \begin{filecontents*}{qwen_math1.5b.csv}
Method,Algebra,Algebra,Algebra,Algebra,Counting & Probability,Counting & Probability,Counting & Probability,Counting & Probability,Geometry,Geometry,Geometry,Geometry,Intermediate Algebra,Intermediate Algebra,Intermediate Algebra,Intermediate Algebra,Number Theory,Number Theory,Number Theory,Number Theory,Prealgebra,Prealgebra,Prealgebra,Prealgebra,Precalculus,Precalculus,Precalculus,Precalculus
Layer,0,5,15,27,0,5,15,27,0,5,15,27,0,5,15,27,0,5,15,27,0,5,15,27,0,5,15,27
Prune Head 0,0.92,0.9,0.91,0.91,0.63,0.63,\textbf{0.79},0.66,0.56,0.49,0.54,0.51,0.59,\textbf{0.62},0.55,\textbf{0.62},0.82,0.87,\textbf{0.89},0.87,\textbf{0.82},0.82,0.79,0.78,0.62,0.59,0.57,0.68
Prune Head 1,0.92,0.92,0.92,0.92,0.71,0.68,0.63,0.63,0.51,0.56,0.59,0.51,0.6,0.52,0.53,0.58,0.84,0.85,0.85,\textbf{0.90},\textbf{0.82},0.77,0.78,0.79,0.62,0.62,0.64,0.66
Prune Head 2,\textbf{0.93},0.91,0.91,0.91,\textbf{0.74},0.66,0.66,\textbf{0.71},\textbf{0.59},0.54,\textbf{0.63},0.54,\textbf{0.61},0.58,0.54,0.59,\textbf{0.92},\textbf{0.89},0.84,0.89,\textbf{0.82},\textbf{0.83},0.77,0.79,0.61,0.62,0.61,0.68
Prune Head 3,0.06,0.91,0.92,0.9,0.08,0.71,0.61,\textbf{0.71},0.07,0.56,0.61,0.49,0.07,0.6,\textbf{0.59},0.6,0.1,0.85,\textbf{0.89},\textbf{0.90},0.12,0.79,\textbf{0.83},0.79,0.09,0.62,0.66,0.68
Prune Head 4,0.89,0.92,\textbf{0.93},0.9,0.71,0.66,0.66,0.68,\textbf{0.59},0.54,0.54,0.51,0.59,0.6,0.55,0.59,0.85,0.82,0.87,0.89,\textbf{0.82},0.78,0.77,0.78,0.62,0.62,0.62,0.68
Prune Head 5,0.56,0.91,0.9,0.91,0.5,0.61,0.61,0.66,0.46,\textbf{0.59},0.54,\textbf{0.56},0.37,0.58,0.53,0.59,0.47,\textbf{0.89},0.81,0.84,0.54,\textbf{0.83},0.76,0.77,0.41,0.64,\textbf{0.68},0.66
Prune Head 6,0.81,0.92,0.9,0.9,0.61,0.66,0.63,0.63,0.56,0.56,0.59,0.51,0.51,0.59,\textbf{0.59},0.57,0.71,0.82,0.82,0.89,0.73,0.82,0.74,0.76,0.54,0.66,0.66,0.64
Prune Head 7,0.91,0.92,0.9,0.9,0.63,0.61,0.63,0.66,0.56,0.56,0.54,0.54,0.58,0.6,0.56,0.59,0.85,0.87,\textbf{0.89},0.87,0.78,0.79,0.8,0.78,0.64,\textbf{0.68},0.55,0.62
Prune Head 8,0.86,\textbf{0.93},0.9,\textbf{0.93},0.63,\textbf{0.74},0.68,0.68,\textbf{0.59},0.49,0.56,0.54,0.53,0.56,\textbf{0.59},0.58,0.82,0.85,0.87,0.85,0.79,0.8,0.82,\textbf{0.80},0.62,0.61,0.62,0.64
Prune Head 9,\textbf{0.93},0.91,0.91,0.91,0.66,0.66,0.68,0.66,0.56,0.49,0.54,0.49,0.57,0.54,0.52,0.57,0.9,0.84,0.85,0.89,0.78,0.82,0.8,0.79,0.64,0.66,0.66,0.66
Prune Head 10,0.92,0.92,0.92,0.89,0.68,0.66,0.68,0.66,0.54,0.51,0.56,0.51,0.55,0.57,0.56,0.57,0.85,0.85,0.84,0.87,0.8,0.8,0.77,0.78,\textbf{0.66},0.62,0.62,\textbf{0.70}
Prune Head 11,0.9,0.92,0.91,0.9,0.68,0.66,0.71,0.66,0.51,0.56,0.56,0.49,0.6,0.61,0.55,0.6,0.87,0.84,0.87,0.87,0.78,0.8,0.77,\textbf{0.80},\textbf{0.66},0.62,0.61,0.66
No Prune,0.91,0.91,0.91,0.91,0.63,0.63,0.63,0.63,0.49,0.49,0.49,0.49,0.58,0.58,0.58,0.57,0.89,\textbf{0.89},\textbf{0.89},0.87,0.79,0.79,0.79,0.79,\textbf{0.66},0.66,0.66,0.68
\end{filecontents*}
\pgfplotstabletypeset[
        col sep=comma,
        string type,
        every head row/.style={
            output empty row,
            before row={
                \toprule
                \multirow{3}{*}{} &
                \multicolumn{4}{c}{Algebra} &
                \multicolumn{4}{c}{Counting \& Probability} &
                \multicolumn{4}{c}{Geometry} &
                \multicolumn{4}{c}{Intermediate Algebra} &
                \multicolumn{4}{c}{Number Theory} &
                \multicolumn{4}{c}{Prealgebra} &
                \multicolumn{4}{c}{Precalculus}\\
                \cmidrule(r){2-5}\cmidrule(r){6-9}\cmidrule(r){10-13}
                \cmidrule(r){14-17}\cmidrule(r){18-21}\cmidrule(r){22-25}
                \cmidrule(r){26-29}\\
            }
        },
        every row no 0/.style={after row=\midrule},
        every last row/.style={after row=\bottomrule},
    ]{qwen_math1.5b.csv} 
    }
    \end{center}
\end{sidewaystable}
\clearpage

\clearpage
\begin{sidewaystable}
    \vspace{60mm}
    \begin{center}
    \caption{We report Pass@1 scores for single head pruning across different layers of \texttt{Qwen2.5-Math-7B-Instruct}, organized by question statement subjects.}
    \label{tab:observe_qwen_math_7}
    \resizebox{\textwidth}{!}{
    \begin{filecontents*}{qwen_math7b.csv}
Method,Algebra,Algebra,Algebra,Algebra,Counting & Probability,Counting & Probability,Counting & Probability,Counting & Probability,Geometry,Geometry,Geometry,Geometry,Intermediate Algebra,Intermediate Algebra,Intermediate Algebra,Intermediate Algebra,Number Theory,Number Theory,Number Theory,Number Theory,Prealgebra,Prealgebra,Prealgebra,Prealgebra,Precalculus,Precalculus,Precalculus,Precalculus
Layer,0,5,15,27,0,5,15,27,0,5,15,27,0,5,15,27,0,5,15,27,0,5,15,27,0,5,15,27
Prune Head 0,0.96,0.96,0.97,\textbf{0.97},\textbf{0.84},0.82,0.84,\textbf{0.84},0.68,0.68,0.59,0.66,\textbf{0.68},0.67,0.64,0.62,0.9,0.95,0.95,\textbf{0.97},0.9,0.88,0.9,0.88,0.71,0.77,0.7,\textbf{0.80}
Prune Head 1,\textbf{0.97},0.96,0.97,\textbf{0.97},\textbf{0.84},0.79,0.84,0.82,0.68,0.66,0.66,0.66,0.63,0.65,0.63,0.62,0.95,0.95,0.97,0.95,0.89,0.89,0.89,0.88,0.73,0.75,0.75,0.77
Prune Head 2,\textbf{0.97},0.96,\textbf{0.98},\textbf{0.97},\textbf{0.84},\textbf{0.84},\textbf{0.87},\textbf{0.84},0.66,0.66,0.66,0.66,0.65,0.64,0.64,0.65,0.92,0.95,0.92,0.95,0.9,0.89,0.9,0.87,0.75,0.73,0.71,0.77
Prune Head 3,0.36,0.97,0.97,\textbf{0.97},0.5,0.82,0.82,0.82,0.29,0.61,0.61,0.66,0.25,0.65,0.68,0.62,0.29,\textbf{0.97},\textbf{1.00},0.95,0.34,0.89,0.88,0.88,0.2,0.77,0.73,0.75
Prune Head 4,\textbf{0.97},0.97,0.95,\textbf{0.97},\textbf{0.84},0.82,0.82,0.82,0.68,0.66,\textbf{0.68},0.66,0.61,0.66,0.63,0.62,0.95,0.94,0.92,0.95,0.89,\textbf{0.90},0.89,\textbf{0.90},0.75,0.73,0.75,0.75
Prune Head 5,0.96,0.95,0.96,\textbf{0.97},0.82,0.82,0.84,0.82,\textbf{0.73},0.63,0.63,0.68,0.63,0.67,0.65,0.62,0.95,0.94,0.97,0.92,0.89,\textbf{0.90},0.88,0.88,0.73,0.75,0.71,0.75
Prune Head 6,\textbf{0.97},0.96,0.96,\textbf{0.97},\textbf{0.84},0.82,0.82,0.82,0.63,0.68,0.63,0.66,0.66,0.61,0.67,0.63,0.92,0.94,0.95,0.95,0.88,0.89,0.9,0.89,0.68,0.73,0.77,0.79
Prune Head 7,0.96,0.97,0.97,\textbf{0.97},0.79,0.82,0.82,0.82,0.61,0.66,0.63,0.73,0.64,0.64,0.66,0.62,0.95,0.95,0.95,0.95,0.89,0.89,0.89,0.89,0.73,\textbf{0.79},0.75,0.71
Prune Head 8,\textbf{0.97},0.97,0.97,\textbf{0.97},0.82,\textbf{0.84},0.84,\textbf{0.84},0.68,0.66,\textbf{0.68},0.68,0.65,0.63,0.65,0.63,\textbf{0.97},0.95,0.94,\textbf{0.97},0.89,0.88,0.88,0.88,0.75,0.75,0.7,0.71
Prune Head 9,0.96,0.97,0.96,\textbf{0.97},\textbf{0.84},\textbf{0.84},0.82,\textbf{0.84},0.61,0.66,0.59,\textbf{0.76},0.63,0.65,0.64,\textbf{0.67},0.94,0.95,0.97,0.95,0.89,0.88,0.9,0.89,0.77,0.75,0.77,0.73
Prune Head 10,0.95,0.97,0.96,0.96,\textbf{0.84},0.82,0.84,\textbf{0.84},0.66,0.63,0.63,0.71,0.64,0.63,0.65,0.64,0.94,0.95,0.94,0.95,0.89,0.88,0.89,0.88,0.73,0.75,0.71,0.75
Prune Head 11,\textbf{0.97},0.97,0.97,\textbf{0.97},0.82,0.82,0.82,\textbf{0.84},0.63,\textbf{0.71},0.63,0.73,0.64,0.65,\textbf{0.69},0.63,\textbf{0.97},0.94,0.95,0.95,0.9,0.88,0.9,0.88,\textbf{0.79},0.71,0.77,0.75
Prune Head 12,\textbf{0.97},0.97,\textbf{0.98},\textbf{0.97},0.82,0.82,\textbf{0.87},\textbf{0.84},0.66,0.66,\textbf{0.68},0.71,0.66,0.63,0.67,0.64,0.95,0.95,0.92,\textbf{0.97},0.88,0.89,0.87,0.88,0.77,0.77,0.75,0.73
Prune Head 13,\textbf{0.97},0.97,\textbf{0.98},\textbf{0.97},0.82,0.82,0.82,\textbf{0.84},0.61,0.68,\textbf{0.68},0.68,0.6,0.64,0.63,0.64,0.95,0.95,0.94,\textbf{0.97},0.88,0.88,\textbf{0.91},0.88,0.73,0.75,\textbf{0.79},0.75
Prune Head 14,0.96,0.97,0.96,\textbf{0.97},\textbf{0.84},0.82,0.84,0.82,0.63,0.61,0.66,0.66,0.67,0.66,0.63,0.61,0.94,0.95,0.95,0.95,0.89,0.88,0.89,0.85,0.77,0.73,0.71,0.75
Prune Head 15,\textbf{0.97},0.97,0.94,\textbf{0.97},0.79,0.82,0.84,0.82,\textbf{0.73},0.68,0.63,0.61,0.66,0.66,0.64,0.63,\textbf{0.97},0.94,0.94,0.95,0.9,0.88,\textbf{0.91},0.88,0.77,0.71,0.75,0.75
Prune Head 16,0.96,0.96,0.97,\textbf{0.97},0.82,\textbf{0.84},0.82,\textbf{0.84},0.68,0.63,\textbf{0.68},0.66,0.67,0.62,0.64,0.63,0.9,0.95,0.97,0.95,0.88,0.89,0.89,0.87,0.73,0.75,0.75,0.75
Prune Head 17,0.96,0.96,\textbf{0.98},\textbf{0.97},0.82,0.82,0.82,0.82,0.66,0.63,\textbf{0.68},0.61,0.65,0.66,0.64,0.63,0.92,0.94,0.92,0.95,0.87,0.88,0.89,0.88,0.75,0.75,0.75,0.75
Prune Head 18,0.96,0.97,0.97,\textbf{0.97},0.82,0.82,\textbf{0.87},0.82,0.61,0.63,0.66,0.66,0.67,0.6,0.61,0.63,0.94,0.94,0.95,0.95,0.89,0.89,0.88,0.88,0.77,0.77,0.77,0.75
Prune Head 19,0.96,0.96,0.97,\textbf{0.97},0.82,0.82,0.79,0.82,0.63,0.68,0.66,0.63,0.67,0.64,0.65,0.64,0.92,0.95,0.95,0.95,0.87,0.89,0.89,\textbf{0.90},0.73,0.71,0.71,0.75
Prune Head 20,0.96,0.97,0.96,\textbf{0.97},\textbf{0.84},0.82,0.82,0.82,0.63,0.61,0.66,0.68,0.65,0.62,0.65,0.62,0.92,0.95,0.94,0.95,0.88,\textbf{0.90},0.89,0.88,0.75,0.71,0.73,0.75
Prune Head 21,\textbf{0.97},0.96,0.97,\textbf{0.97},0.79,\textbf{0.84},0.82,\textbf{0.84},0.63,0.59,0.66,0.66,0.64,0.62,0.66,0.61,0.9,0.95,0.95,0.95,\textbf{0.91},0.88,0.89,0.88,0.73,0.75,0.71,0.71
Prune Head 22,\textbf{0.97},0.97,0.96,\textbf{0.97},\textbf{0.84},\textbf{0.84},0.82,\textbf{0.84},0.71,0.66,\textbf{0.68},0.66,0.62,0.63,0.62,0.61,0.95,0.95,0.97,0.94,0.9,0.89,0.88,0.88,0.73,0.73,0.75,0.75
Prune Head 23,0.96,0.97,\textbf{0.98},\textbf{0.97},\textbf{0.84},\textbf{0.84},0.82,0.82,0.63,0.66,0.61,0.63,0.65,0.64,0.65,0.64,0.95,0.94,0.94,0.95,0.9,0.89,0.9,0.88,0.73,0.73,0.77,0.75
Prune Head 24,\textbf{0.97},0.96,0.96,\textbf{0.97},0.82,0.82,0.79,\textbf{0.84},\textbf{0.73},0.63,0.66,0.63,0.65,0.62,0.66,0.62,0.94,0.95,0.9,0.95,0.89,0.88,0.9,0.89,0.73,0.77,0.77,0.77
Prune Head 25,\textbf{0.97},0.97,0.96,\textbf{0.97},\textbf{0.84},0.82,0.79,0.82,0.63,\textbf{0.71},\textbf{0.68},0.66,0.64,\textbf{0.68},0.61,0.61,0.95,0.94,0.94,0.95,0.88,0.89,\textbf{0.91},\textbf{0.90},0.75,0.73,0.73,0.77
Prune Head 26,\textbf{0.97},\textbf{0.98},\textbf{0.98},\textbf{0.97},0.79,0.82,0.79,\textbf{0.84},0.71,0.68,\textbf{0.68},0.68,0.62,0.67,0.62,0.61,0.95,0.95,0.95,0.94,0.88,0.88,0.88,0.89,0.73,\textbf{0.79},0.75,0.77
Prune Head 27,\textbf{0.97},0.97,0.97,\textbf{0.97},\textbf{0.84},0.79,0.82,0.82,0.66,0.66,0.66,0.61,0.62,0.65,0.64,0.63,0.95,0.95,0.95,0.94,0.88,\textbf{0.90},\textbf{0.91},0.88,0.73,0.73,0.68,0.75
No Prune,\textbf{0.97},0.97,0.97,\textbf{0.97},0.82,0.82,0.82,0.82,0.66,0.66,0.66,0.66,0.62,0.62,0.62,0.62,0.95,0.95,0.95,0.95,0.88,0.88,0.88,0.88,0.75,0.75,0.75,0.75
\end{filecontents*}
\pgfplotstabletypeset[
        col sep=comma,
        string type,
        every head row/.style={
            output empty row,
            before row={
                \toprule
                \multirow{3}{*}{} &
                \multicolumn{4}{c}{Algebra} &
                \multicolumn{4}{c}{Counting \& Probability} &
                \multicolumn{4}{c}{Geometry} &
                \multicolumn{4}{c}{Intermediate Algebra} &
                \multicolumn{4}{c}{Number Theory} &
                \multicolumn{4}{c}{Prealgebra} &
                \multicolumn{4}{c}{Precalculus}\\
                \cmidrule(r){2-5}\cmidrule(r){6-9}\cmidrule(r){10-13}
                \cmidrule(r){14-17}\cmidrule(r){18-21}\cmidrule(r){22-25}
                \cmidrule(r){26-29}\\
            }
        },
        every row no 0/.style={after row=\midrule},
        every last row/.style={after row=\bottomrule},
    ]{qwen_math7b.csv} 
    }
    \end{center}
\end{sidewaystable}
\clearpage

\clearpage
\begin{sidewaystable}
    \vspace{60mm}
    \begin{center}
    \caption{We report Pass@1 scores for single head pruning across different layers of \texttt{Qwen2.5-7B-Instruct}, organized by question statement subjects.}
    \label{tab:observe_qwen_7}
    \resizebox{\textwidth}{!}{
    \begin{filecontents*}{qwen7b.csv}
Method,Algebra,Algebra,Algebra,Algebra,Counting & Probability,Counting & Probability,Counting & Probability,Counting & Probability,Geometry,Geometry,Geometry,Geometry,Intermediate Algebra,Intermediate Algebra,Intermediate Algebra,Intermediate Algebra,Number Theory,Number Theory,Number Theory,Number Theory,Prealgebra,Prealgebra,Prealgebra,Prealgebra,Precalculus,Precalculus,Precalculus,Precalculus
Layer,0,5,15,27,0,5,15,27,0,5,15,27,0,5,15,27,0,5,15,27,0,5,15,27,0,5,15,27
Prune Head 0,0.55,0.94,0.95,0.95,0.5,0.74,\textbf{0.76},0.71,0.44,0.56,0.56,0.59,0.38,0.55,0.54,0.57,0.34,0.85,0.85,0.87,0.49,0.83,0.84,0.87,0.36,0.59,0.61,0.59
Prune Head 1,0.85,0.94,0.93,0.94,0.61,0.71,0.74,0.71,0.49,0.63,0.56,0.61,0.51,0.57,0.53,0.59,0.82,0.87,0.87,0.87,0.73,0.84,0.83,0.87,\textbf{0.70},0.62,0.59,0.61
Prune Head 2,0.94,0.94,0.95,0.94,0.68,0.71,0.71,0.71,0.63,0.61,\textbf{0.71},\textbf{0.66},0.58,0.56,0.54,0.58,0.82,0.84,0.81,0.85,0.87,0.84,0.83,0.87,0.61,0.61,0.61,0.64
Prune Head 3,0.02,0.94,0.92,0.94,0,\textbf{0.76},\textbf{0.76},0.74,0.05,0.61,0.68,0.61,0.04,0.58,\textbf{0.60},0.6,0,0.85,0.84,0.85,0.01,0.84,0.84,0.87,0.02,0.61,0.62,0.61
Prune Head 4,0.94,0.94,0.94,0.94,0.71,0.68,0.74,0.71,0.61,0.59,0.56,0.61,0.59,0.58,0.54,0.59,\textbf{0.85},0.82,0.79,0.85,0.87,0.84,0.83,0.85,0.62,0.57,0.64,0.61
Prune Head 5,0.95,0.95,0.93,0.94,0.71,0.74,0.74,0.71,0.59,0.61,0.61,0.56,0.58,0.54,0.57,0.6,\textbf{0.85},0.84,0.84,0.82,0.87,0.82,0.84,0.85,0.62,0.62,0.62,0.61
Prune Head 6,0.63,0.95,0.91,0.95,0.53,\textbf{0.76},0.66,0.71,0.54,0.59,0.56,0.61,0.44,0.57,0.53,0.6,0.45,\textbf{0.89},0.84,0.87,0.65,0.85,0.83,0.85,0.39,0.62,0.64,0.62
Prune Head 7,\textbf{0.96},0.93,0.91,0.94,\textbf{0.76},0.74,\textbf{0.76},0.74,\textbf{0.68},0.61,0.54,0.59,0.55,0.57,0.59,0.58,0.84,0.87,0.82,0.87,0.84,0.85,0.85,0.84,0.64,0.64,0.61,0.64
Prune Head 8,0.94,0.95,0.94,0.94,0.74,0.74,0.74,0.74,0.63,0.63,0.59,0.61,0.58,0.58,0.58,0.58,0.82,0.87,0.84,0.84,0.85,0.85,0.85,0.85,0.55,0.57,0.59,0.59
Prune Head 9,0.95,0.95,0.94,\textbf{0.97},0.68,0.74,0.68,0.71,0.56,0.66,0.56,0.59,0.58,0.55,0.57,0.57,\textbf{0.85},0.85,\textbf{0.92},0.87,0.87,0.87,\textbf{0.87},0.85,0.61,0.57,0.62,0.64
Prune Head 10,0.91,0.94,0.94,0.94,0.61,0.74,0.68,0.76,0.54,0.61,0.63,0.61,\textbf{0.60},0.6,0.55,0.56,0.82,0.85,0.77,0.85,0.82,0.85,0.8,0.8,0.61,0.61,0.64,0.55
Prune Head 11,0.87,0.93,0.95,0.94,0.71,0.74,\textbf{0.76},0.74,0.59,0.61,0.56,0.61,0.57,0.56,0.55,0.57,0.77,0.87,0.89,0.82,0.83,0.84,0.84,0.84,0.52,0.62,0.68,0.61
Prune Head 12,0.94,0.94,0.94,\textbf{0.97},0.74,0.71,0.74,0.74,0.63,0.59,0.59,0.61,0.56,0.59,0.57,0.57,0.84,0.85,0.89,0.87,0.87,0.84,0.82,0.82,0.68,0.61,0.57,0.57
Prune Head 13,0.95,0.95,0.93,0.94,0.71,0.71,0.71,\textbf{0.79},0.61,0.63,0.61,0.59,0.57,0.6,\textbf{0.60},0.54,\textbf{0.85},\textbf{0.89},0.87,0.85,0.87,0.88,0.83,0.83,0.62,0.59,0.55,0.55
Prune Head 14,0.94,0.93,0.94,0.94,0.71,0.71,0.68,0.76,0.63,0.56,0.61,0.59,0.54,0.55,0.55,0.58,0.77,0.84,0.85,0.84,0.82,0.85,0.84,0.87,\textbf{0.70},0.61,0.64,0.61
Prune Head 15,0.81,0.94,0.94,0.95,0.68,0.74,\textbf{0.76},0.71,0.54,0.56,0.61,0.61,0.52,0.56,0.58,0.59,0.77,0.84,0.85,0.82,0.82,0.88,0.85,\textbf{0.88},0.59,0.57,0.64,0.59
Prune Head 16,\textbf{0.96},0.94,0.94,0.94,0.68,0.68,\textbf{0.76},0.74,0.66,0.61,0.63,0.59,0.58,0.61,0.54,0.59,0.84,0.82,0.87,0.84,0.88,0.87,0.82,\textbf{0.88},0.59,0.62,0.62,0.62
Prune Head 17,0.93,0.95,0.94,0.94,0.71,0.71,0.74,0.76,0.59,0.59,0.63,0.59,0.54,0.55,\textbf{0.60},\textbf{0.61},0.81,0.84,0.87,0.85,\textbf{0.89},0.87,0.82,0.82,0.62,0.59,0.61,0.61
Prune Head 18,0.94,0.94,0.94,0.94,0.74,0.71,0.74,0.71,0.66,0.61,0.56,0.61,0.56,0.56,0.58,0.56,\textbf{0.85},0.84,0.85,0.87,0.87,0.85,0.83,0.84,0.64,0.59,0.66,0.61
Prune Head 19,0.94,\textbf{0.96},0.92,0.94,0.71,\textbf{0.76},\textbf{0.76},0.71,0.63,0.56,0.66,0.59,0.56,0.55,0.53,0.58,0.82,0.87,0.84,0.85,0.87,0.79,\textbf{0.87},0.85,0.66,0.57,\textbf{0.70},0.61
Prune Head 20,0.95,0.94,0.94,0.94,0.74,0.74,0.74,0.71,0.66,0.59,0.59,0.61,\textbf{0.60},0.57,0.57,0.58,0.81,0.87,0.82,0.84,0.84,0.82,0.84,\textbf{0.88},0.61,0.62,0.66,0.61
Prune Head 21,0.95,\textbf{0.96},\textbf{0.96},0.96,\textbf{0.76},0.74,0.74,0.71,0.61,0.61,0.61,0.61,0.56,0.58,0.58,0.6,\textbf{0.85},0.85,0.85,0.87,0.85,0.85,0.84,0.87,0.61,0.62,0.62,0.61
Prune Head 22,0.44,\textbf{0.96},0.93,0.94,0.24,\textbf{0.76},0.74,0.71,0.22,0.59,0.66,0.59,0.12,0.58,0.55,\textbf{0.61},0.26,0.85,0.87,0.84,0.52,\textbf{0.89},0.83,0.84,0.07,0.62,0.64,0.61
Prune Head 23,0.11,0.94,0.92,0.94,0.21,0.74,0.63,0.71,0.17,0.66,0.68,0.59,0.04,\textbf{0.62},0.49,0.58,0.06,0.84,0.79,0.87,0.27,0.85,0.82,0.87,0.04,0.61,0.59,0.59
Prune Head 24,0.83,0.94,0.93,0.96,0.61,0.74,0.71,0.71,0.56,0.66,0.63,\textbf{0.66},0.51,0.6,0.56,0.6,0.77,0.85,0.84,0.85,0.68,0.87,0.84,0.84,0.57,0.62,0.55,0.62
Prune Head 25,0.02,0.95,0.95,0.95,0,\textbf{0.76},0.74,0.66,0,0.63,0.66,0.63,0,0.55,\textbf{0.60},0.59,0,\textbf{0.89},0.81,0.85,0.02,0.84,0.83,0.85,0.02,\textbf{0.70},0.59,\textbf{0.66}
Prune Head 26,0.5,0.95,0.93,0.95,0.47,0.71,0.71,0.71,0.37,0.54,0.54,0.56,0.25,0.58,\textbf{0.60},0.56,0.56,0.85,0.89,\textbf{0.89},0.63,0.88,0.85,0.85,0.21,0.55,0.64,\textbf{0.66}
Prune Head 27,0.83,\textbf{0.96},0.94,0.95,0.68,0.71,0.71,0.63,0.46,\textbf{0.68},0.59,0.56,0.49,0.6,0.52,0.59,0.69,0.85,0.82,0.82,0.63,0.84,0.84,0.85,0.61,0.59,0.59,0.64
No Prune,0.94,0.94,0.94,0.94,0.74,0.74,0.74,0.74,0.59,0.59,0.59,0.59,0.59,0.59,0.59,0.59,0.84,0.84,0.84,0.84,0.87,0.87,\textbf{0.87},0.87,0.59,0.59,0.59,0.59
\end{filecontents*}
\pgfplotstabletypeset[
        col sep=comma,
        string type,
        every head row/.style={
            output empty row,
            before row={
                \toprule
                \multirow{3}{*}{} &
                \multicolumn{4}{c}{Algebra} &
                \multicolumn{4}{c}{Counting \& Probability} &
                \multicolumn{4}{c}{Geometry} &
                \multicolumn{4}{c}{Intermediate Algebra} &
                \multicolumn{4}{c}{Number Theory} &
                \multicolumn{4}{c}{Prealgebra} &
                \multicolumn{4}{c}{Precalculus}\\
                \cmidrule(r){2-5}\cmidrule(r){6-9}\cmidrule(r){10-13}
                \cmidrule(r){14-17}\cmidrule(r){18-21}\cmidrule(r){22-25}
                \cmidrule(r){26-29}\\
            }
        },
        every row no 0/.style={after row=\midrule},
        every last row/.style={after row=\bottomrule},
    ]{qwen7b.csv} 
    }
    \end{center}
\end{sidewaystable}
\clearpage

\end{document}